\pdfoutput=1
\documentclass[11pt]{article}
\usepackage[preprint]{acl}
\usepackage{times}
\usepackage{latexsym}
\usepackage[T1]{fontenc}
\usepackage[utf8]{inputenc}
\usepackage{microtype}
\usepackage{inconsolata}
\usepackage{amsmath}
\usepackage{mathtools}
\usepackage{bm}
\usepackage{amssymb}
\usepackage{graphicx}
\usepackage{graphicx,xcolor}
\usepackage{booktabs}
\usepackage {colortbl,array}
\usepackage{multirow}
\usepackage{enumitem}
\usepackage{tabularx}
\usepackage{ragged2e}

\title{
  \centering
  \makebox[\textwidth][c]{
    \raisebox{-0.2ex}{\includegraphics[height=2.3em]{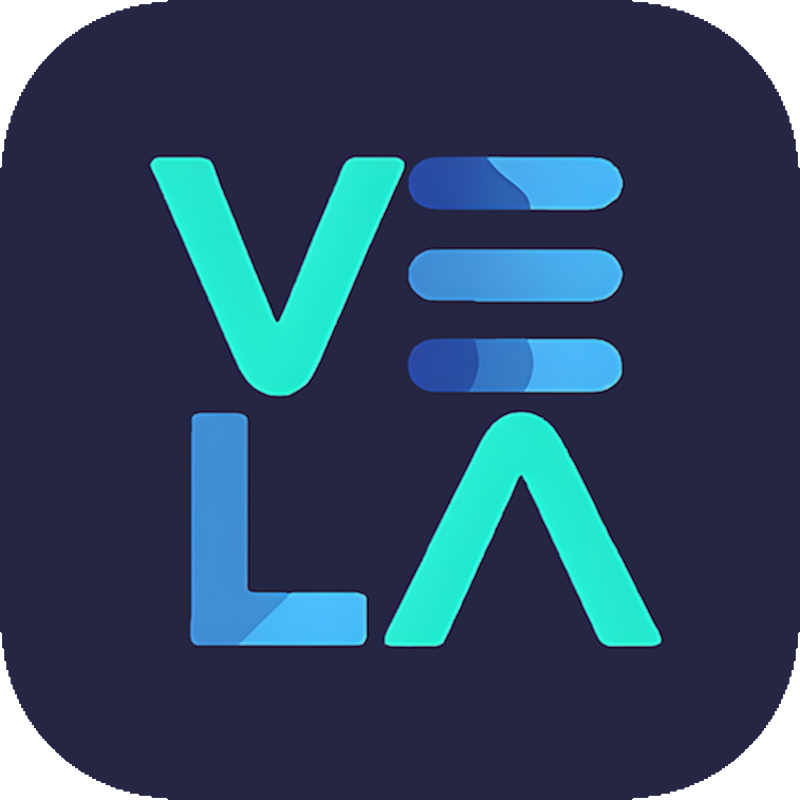}}
    \hspace{-2.5em}
    \parbox[b]{0.8\textwidth}{\centering
      \textbf{\Large \textsc{Vela}: An LLM-Hybrid-as-a-Judge Approach}\\
      \textbf{\Large for Evaluating Long Image Captions}
    }
  }
}

\author{
  Kazuki Matsuda \;
  Yuiga Wada \;
  Shinnosuke Hirano \;
  Seitaro Otsuki \;
  Komei Sugiura \\
  Keio University \\
  \texttt{\{k2matsuda0, yuiga, shinhirano, otsu8sei14, komei.sugiura\}@keio.jp}
}

\begin{document}
\maketitle

\begin{abstract}
In this study, we focus on the automatic evaluation of long and detailed image captions generated by multimodal Large Language Models (MLLMs).
Most existing automatic evaluation metrics for image captioning are primarily designed for short captions and are not suitable for evaluating long captions.
Moreover, recent LLM-as-a-Judge approaches suffer from slow inference due to their reliance on autoregressive inference and early fusion of visual information.
To address these limitations, we propose \textsc{Vela}, an automatic evaluation metric for long captions developed within a novel LLM-Hybrid-as-a-Judge framework.
Furthermore, we propose LongCap-Arena, a benchmark specifically designed for evaluating metrics for long captions.
This benchmark comprises 7,805 images, the corresponding human-provided long reference captions and long candidate captions, and 32,246 human judgments from three distinct perspectives: \textit{Descriptiveness}, \textit{Relevance}, and \textit{Fluency}.
We demonstrated that \textsc{Vela} outperformed existing metrics and achieved superhuman performance on LongCap-Arena.
Our code and dataset are available at \url{https://vela.kinsta.page/}.
\end{abstract}
\section{Introduction}

\begin{figure}[t]
    \centering
    \includegraphics[width=1.0\linewidth]{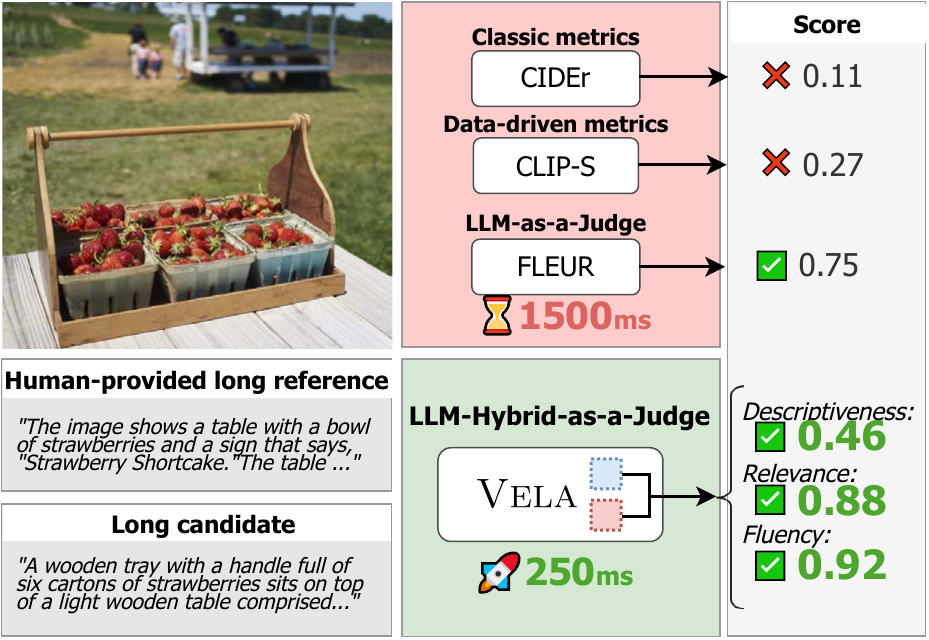}
    \caption{Overview of \textsc{Vela}, which evaluates long image captions from three perspectives: \textit{Descriptiveness}, \textit{Relevance}, and \textit{Fluency}. \textsc{Vela} employs an LLM-Hybrid-as-a-Judge framework, which enables both computational efficiency and high alignment with human judgments.}
    \vspace{-3mm}
    \label{fig:eye-catch}
\end{figure}
Multimodal Large Language Models (MLLMs) have been widely researched and applied in various social domains, including robotics and healthcare.~\cite{gpt4, gemini, llava, llava-1.5, vila, instructblip, qwen-vl}.
To effectively develop MLLMs, it is essential to use automatic evaluation metrics that closely align with human judgments.
Despite the proficiency of MLLMs in generating long and detailed captions, effective evaluation metrics for assessing this capability have not yet been fully established.
Indeed, classic metrics, such as BLEU~\cite{bleu} and CIDEr~\cite{cider}, have been shown to exhibit weak correlation with human judgments when evaluating long captions.

In this study, we focus on the automatic evaluation of long and detailed image captions generated by MLLMs.
Our target task is challenging even for humans, as we will demonstrate in Section~\ref{quantitative}.

Although there have been many attempts to develop evaluation metrics for image captions, they remain inadequate for evaluating long captions.
Indeed, recent approaches~\cite{clipscore, pac-s, pac-spp, polos, deneb} exhibit low correlation with human judgments.
Moreover, LLM-as-a-Judge approaches~\cite{chan2023clair, lee2024fleur, tong2025gveval, yao2024hifi} are impractical because of their slow inference.
In fact, they require over three hours to evaluate generated captions on standard benchmarks (e.g.,~\cite{coco, nocaps}).
This is largely attributed to the autoregressive nature of LLM-based inference and the early fusion of visual information.

To address these limitations, we propose \textsc{Vela}, an automatic evaluation metric for long image captions, developed within a novel LLM-Hybrid-as-a-Judge framework.
Fig.~\ref{fig:eye-catch} presents an overview of \textsc{Vela} with a typical sample.
To train and validate the proposed metric, we construct the LongCap-Arena benchmark, which includes images, human-provided long reference captions, long candidate captions, and human judgments.

\textsc{Vela} distinguishes itself from existing metrics in two key aspects:
First, \textsc{Vela} adopts a late fusion approach to integrate visual information with a non-autoregressive LLM, unlike existing metrics (e.g. \cite{lee2024fleur, tong2025gveval}).
This late fusion approach avoids increases in the input sequence lengths, enabling inference that is faster than that of early fusion approaches.
Second, instead of outputting a single score that represents the overall quality of a candidate, the proposed metric outputs evaluation scores across three distinct perspectives: \textit{Descriptiveness} (\textit{Desc.}), \textit{Relevance} (\textit{Rel.}), and \textit{Fluency} (\textit{Flu.})
This prevents certain evaluation criteria from being ignored, which is a common issue in metrics that output only a single score~\cite{ohi2024harmoniceval}.

The main contributions of this study are summarized as follows:
\begin{enumerate}
    \setlength{\parskip}{0.5mm}
    \setlength{\itemsep}{0.2mm}
    \item We propose \textsc{Vela}, a supervised metric evaluating long image captions from three distinct perspectives.
    \item We introduce an LLM-Hybrid-as-a-Judge framework, which enables computationally efficient and LLM-based evaluations while incorporating images through a Reference-to-Candidate LLM (R2C-LLM) branch and an Image-to-Candidate Alignment (I2C-Align) branch.
    \item We construct LongCap-Arena, a benchmark for both training and evaluating metrics on long captions, featuring 32,246 human judgments collected from 1,020 annotators.
    \item \textsc{Vela} outperformed existing metrics, including LLM-as-a-Judge approaches, and achieved superhuman performance on the LongCap-Arena benchmark.
\end{enumerate}

\section{Related Work}
Several survey papers on MLLMs and evaluation for image captioning~\cite{ic-survey, ghandi2022deep, mllm-revolution, survey-llm-judge} provide comprehensive overviews of standard models and automatic evaluation metrics.
In particular, \cite{survey-llm-judge} provided a broad summary of LLM-as-a-Judge approaches across various text generation tasks, including image captioning.

\paragraph{Image captioning metrics.}
Standard automatic evaluation metrics for image captioning include BLEU~\cite{bleu}, METEOR~\cite{meteor}, ROUGE~\cite{rouge}, CIDEr~\cite{cider}, and SPICE~\cite{spice}.
Extensions such as CIDEr-R~\cite{cider-r} and JaSPICE~\cite{jaspice} have been proposed.
Although these classic metrics have been widely used in image captioning, researchers have shown that they weakly correlate with human judgments~\cite{clipscore, pac-s, pac-spp, polos, deneb}.

As a result, data-driven evaluation metrics~\cite{vilbertscore,umic,clipscore,pac-s,pac-spp,polos,deneb} have been proposed that leverage pretrained models.
However, these metrics primarily target short captions and are not suitable for evaluating long captions.
HiFiScore~\cite{yao2024hifi} is one of the few metrics targeting long captions, which transforms both images and candidate captions into hierarchical parsing graphs and evaluates the candidates based on node-level matching between the corresponding graphs.
Although it performs well on short captions (e.g.,~\cite{composite, flickr}), converting both the caption and image into hierarchical parsing graphs can lead to information loss when evaluating long captions.

\paragraph{LLM-as-a-Judge approaches.}
Automatic evaluation metrics based on LLMs or MLLMs, often referred to as LLM-as-a-Judge approaches, have been shown to be successful across a variety of evaluation tasks~\cite{survey-llm-judge}.
Several LLM-as-a-Judge approaches have also been proposed for the automatic evaluation of image captioning.
For example, FLEUR~\cite{lee2024fleur} uses an MLLM (LLaVA~\cite{llava, llava-1.5}) and performs early fusion of the visual inputs to enable evaluation with an image input.
Similarly, G-VEval~\cite{tong2025gveval} and HarmonicEval~\cite{ohi2024harmoniceval} also employ MLLMs to incorporate visual information and provide more interpretable evaluations by scoring captions from multiple perspectives.
Despite their advantages, MLLM-based metrics often suffer from slow inference, which results from both the increase in input sequence length caused by early fusion of the visual inputs and the use of autoregressive inference.
Indeed, these metrics require over three hours to evaluate generated captions on standard benchmarks (e.g., COCO~\cite{coco}, nocaps~\cite{nocaps}).
Such inefficiency leads to practical issues in the use of these metrics during the development of MLLMs.

\paragraph{Datasets and benchmarks.}
Standard datasets for evaluating image captioning metrics include Composite~\cite{composite}, Flickr8K-Expert, Flickr8K-CF~\cite{flickr}, Polaris~\cite{polos}, and Nebula~\cite{deneb}.
However, these datasets provide human judgments from a single evaluation perspective only, limiting their ability to capture the diverse quality dimensions of the candidates.

Several recent studies have proposed datasets that include multi-dimensional human judgments~\cite{ohi2024harmoniceval, thumb}.
MMHE~\cite{ohi2024harmoniceval} provides 4,500 human judgments for 100 images across five evaluation perspectives: \textit{Correctness}, \textit{Completeness}, \textit{Clarity}, \textit{Fluency}, and \textit{Conciseness}.  
Similarly, THumB~\cite{thumb} includes 2,500 human judgments for 500 images, across two dimensions: \textit{Precision} and \textit{Recall}.
However, all these datasets are limited to short captions; Composite, Flickr8K-CF, Polaris, and THumB have average caption lengths of 12.6, 11.4, 9.4, and 10.2 words, respectively.
Although the number of datasets for evaluating long captions remains limited, ParaEval~\cite{yao2024hifi} is a representative example.
ParaEval is based on the ImageParagraph dataset~\cite{imageparagraph}, which contains 4,000 images paired with long references.
For each reference, it provides automatically generated negative samples based on four error types: \textit{plausible}, \textit{attribute}, \textit{object}, and \textit{relation}.
 
\section{Problem Statement}
\paragraph{Automatic evaluation for long captions.}
\begin{figure}[t]
    \centering
    \includegraphics[width=1.0\linewidth]{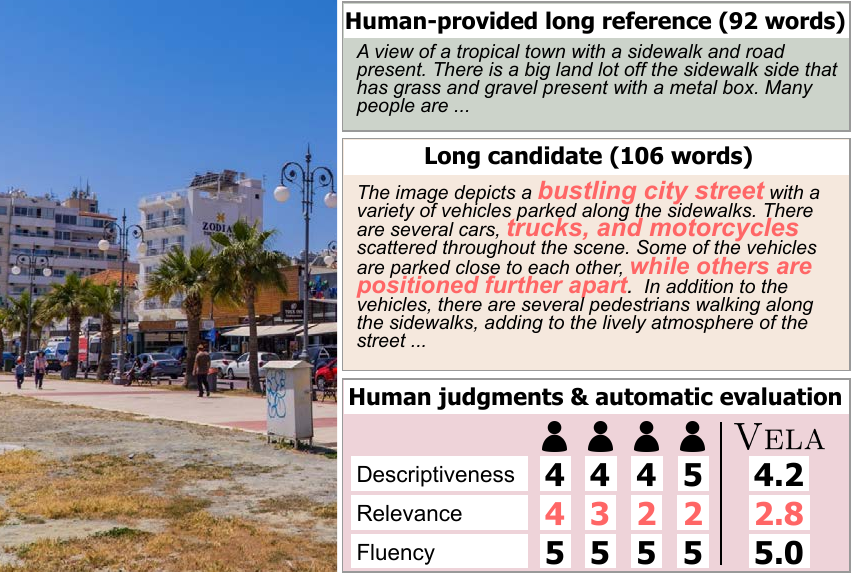}
    \caption{Example of automatic evaluation for long captions.
    In this task, automatic evaluation metrics assess a candidate based on the given image and human-provided long references across three perspectives: \textit{Descriptiveness}, \textit{Relevance}, and \textit{Fluency}.
    The evaluation scores should align with human judgments.}
    \label{fig:task}
    \vspace{-3mm}
\end{figure}
We focus on the automatic evaluation of long and detailed image captions generated by MLLMs.
Fig.~\ref{fig:task} illustrates an automatic evaluation of long captions.
In this task, given an image $\bm{x}_\mathrm{img}$, a long candidate $\bm{x}_\mathrm{cand}$, and $N$ human-provided long references $\{\bm{x}_{\text{ref}}^{(i)}\}_{i=1}^{N}$, automatic evaluation metrics assess $\bm{x}_\mathrm{cand}$ in relation to both $\bm{x}_\mathrm{img}$ and $\{\bm{x}_{\text{ref}}^{(i)}\}_{i=1}^{N}$ from three perspectives.
It is desirable for the metrics to output scores that closely align with human judgments.

These three perspectives are defined as follows:
\begin{itemize}[leftmargin=10pt, itemsep=0pt]
  \item \textbf{\textit{Descriptiveness (Desc.)}} evaluates how thoroughly the candidate caption $\bm{x}_\mathrm{cand}$ captures the details of the image $\bm{x}_\mathrm{img}$, including objects, attributes, and relationships.
  \item \textbf{\textit{Relevance (Rel.)}} measures the extent to which $\bm{x}_\mathrm{cand}$ appropriately reflects the content of $\bm{x}_\mathrm{img}$, by identifying errors such as incorrect objects (e.g., ``dog'' instead of ``cat''), attributes (e.g., ``blue'' instead of ``red''), and relationships (e.g., ``under'' instead of ``on top of'').
  \item \textbf{\textit{Fluency (Flu.)}} focuses on the grammatical correctness, coherence, and naturalness of $\bm{x}_\mathrm{cand}$, including any grammatical or spelling errors, redundancy, and unnecessary phrases that affect linguistic quality.
\end{itemize}
We identified these three perspectives based on the conventions of various natural language generation tasks such as image captioning~\cite{umic, composite, thumb, yao2024hifi, liu2019iccv, yue2023nips}, text summarization~\cite{emnlp19kryscinski, summeval, acl24song}, and machine translation~\cite{tacl21freitag}.
 
\section{Method}
\begin{figure}[t]
    \centering
    \includegraphics[width=0.95\linewidth]{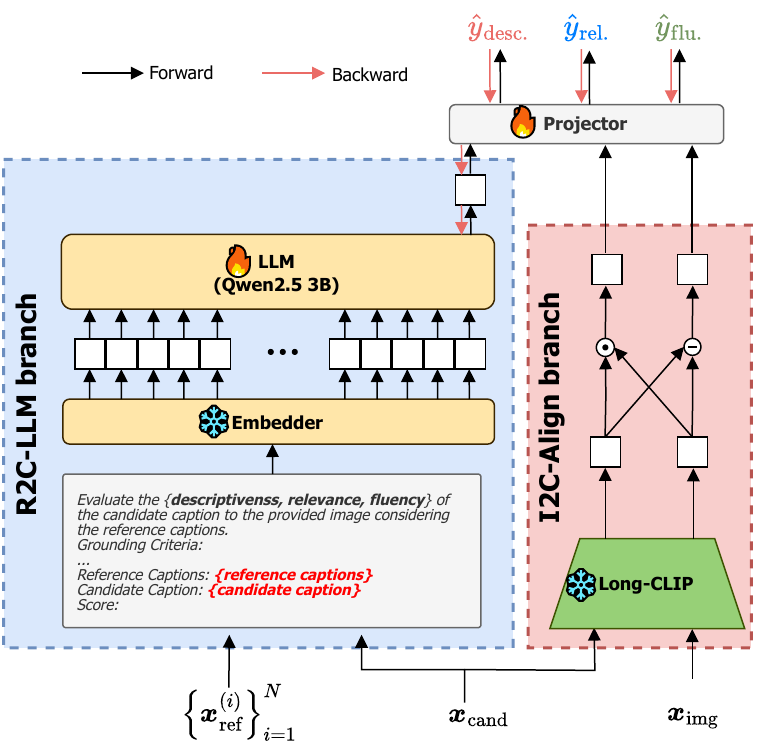}
    \caption{Architecture of \textsc{Vela}. The image, long candidate, and human-provided long references are processed by our metric through two branches: R2C-LLM and I2C-Align. The R2C-LLM branch leverages an LLM to capture the linguistic relationship between the candidate and references, whereas the I2C-Align branch uses Long-CLIP to compute the similarity between the candidate and image.
}
    \label{fig:model}
\end{figure}

We propose \textsc{Vela}, an automatic evaluation metric tailored for evaluating long and detailed image captions. 
Fig.~\ref{fig:model} shows the architecture of \textsc{Vela}.
It consists of two main branches: the R2C-LLM branch and I2C-Align branch.

\subsection{R2C-LLM branch.}
This branch efficiently assesses the quality of $\bm{x}_\mathrm{cand}$ in relation to the corresponding $\bm{x}_\mathrm{ref}$ by employing a lightweight LLM in a non-autoregressive manner.  
We adopt evaluation based on LLMs to take advantage of their extensive world knowledge and linguistic capability acquired through pretraining on broad-domain datasets.
To address the slow speed of autoregressive inference in LLM-as-a-Judge, we employ a non-autoregressive approach that significantly reduces the inference time.  
Furthermore, although MLLMs typically perform early fusion of visual information, which results in increased computational costs and slow inference, we opt for a text-only LLM with a late fusion approach to mitigate these issues.

In the R2C-LLM branch, we first prepare a prompt $\bm{x}_\mathrm{prompt}$ using $\bm{x}_\mathrm{cand}$ and $\{ \bm{x}_\mathrm{ref}^{(i)} \}_{i=1}^N$.  
Our evaluation prompt, designed based on previous work~\cite{flickr, summeval, tong2025gveval, umic}, is provided in Appendix~\ref{sec:prompts}.
We then feed $\bm{x}_\mathrm{prompt}$ into a text-only LLM (Qwen2.5-3B~\cite{qwen2.5}), and obtain the last hidden states in a non-autoregressive manner.  
The sequence of hidden states is denoted as $\{ \bm{h}_i \}_{i=1}^M$, where $M$ denotes the sequence length.
Similar to previous works using the last hidden states (e.g., ~\cite{llm2vec, one, repet}), we compute $\bm{g}_{\text{r2c}}$, the output of the R2C-LLM branch, as follows:
\vspace{-2mm}
\[
\bm{g}_{\text{r2c}} = \left[ \frac{1}{M} \sum_{i=1}^{M} \bm{h}_i \; , \; \bm{h}_M \right].
\]

\subsection{I2C-Align branch.}
This branch evaluates $\bm{x}_\mathrm{cand}$ with respect to $\bm{x}_\mathrm{img}$ using Long-CLIP~\cite{long-clip} without relying on MLLMs.  
As previously mentioned, the early fusion of visual information in MLLM-based metrics results in high computational costs~\cite{chan2023clair, lee2024fleur, tong2025gveval}.  
To avoid these costs, the I2C-Align branch does not employ MLLMs.

The I2C-Align branch uses Long-CLIP to extract $\bm{h}_\text{img}$ and $\bm{h}_\text{cand}$ from $\bm{x}_\text{img}$ and $\bm{x}_\text{cand}$, respectively.  
Unlike existing metrics based on CLIP~\cite{clipscore, pac-s, pac-spp, bridge, polos, deneb}, the I2C-Align branch employs Long-CLIP to overcome the 77-token limit of the original CLIP model, which is insufficient for processing long captions that typically exceed 100 words.

The output of I2C-Align ($\bm{g}_\text{i2c}$) is then computed as follows:
\[
\bm{g}_\text{i2c} = \left[ \left| \bm{h}_\text{img} - \bm{h}_\text{cand} \right| \; , \; \bm{h}_\text{img} \odot \bm{h}_\text{cand} \right],
\]
where $\left| \bm{h}_\text{img} - \bm{h}_\text{cand} \right|$ and $\bm{h}_\text{img} \odot \bm{h}_\text{cand}$ denote the absolute element-wise difference and Hadamard product between $\bm{h}_\text{img}$ and $\bm{h}_\text{cand}$, respectively.  
These operations have been shown to be effective in automatic evaluation across various text generation tasks, such as machine translation and image captioning~\cite{ruse, comet, polos, deneb}.

The final scores $\hat{\bm{y}} \in \mathbb{R}^3$ are computed as follows:
\[
\hat{\bm{y}} = (\hat{y}_{\text{desc}}, \hat{y}_{\text{rel}}, \hat{y}_{\text{flu}}) = \sigma\left( \bm{W} \left[ \bm{g}_{\text{r2c}} , \bm{g}_{\text{i2c}} \right] + \bm{b} \right),
\]
where $\sigma$ denotes the sigmoid function, and $\bm{W}$ and $\bm{b}$ are trainable parameters.  
Here, $\hat{y}_{\text{desc}}$, $\hat{y}_{\text{rel}}$, and $\hat{y}_{\text{flu}}$ denote the predicted scores for \textit{Desc.}, \textit{Rel.}, and \textit{Flu.}, respectively.
We employed the mean squared error as our loss function.

\section{Experiments}
\subsection{Experimental Setup}
\paragraph{LongCap-Arena benchmark.}
We constructed LongCap-Arena, a benchmark specifically designed to evaluate metrics for long image captions. 

To the best of our knowledge, few datasets specifically focus on evaluating metrics for long captions~\cite{yao2024hifi}. 
Existing long-caption datasets for metric evaluation (e.g., ParaEval\cite{yao2024hifi}) do not contain human-provided references annotated based on the semantic structures of images~\cite{dci, imageparagraph, localized}.
Moreover, candidates in ParaEval are limited to either human-provided references or negative examples generated through simple word replacements, limiting diversity in candidate quality.

To address these limitations, we constructed LongCap-Arena, a benchmark that enables comprehensive and authorized evaluation by providing candidate captions with diverse quality and human-provided long references derived from the DCI dataset~\cite{dci}.
This benchmark comprises images, long candidate captions, human-provided long reference captions, and human judgments obtained by assessing long candidate captions from three perspectives.
Unlike existing datasets with human judgments~\cite{composite, flickr, polos}, LongCap-Arena contains captions with over 100 words on average.
Moreover, LongCap-Arena provides human judgments from multiple perspectives, in contrast to most existing image captioning datasets, which assess only the overall appropriateness of the candidates.

To construct LongCap-Arena, we used images and long reference captions from the DCI dataset~\cite{dci}. 
The DCI dataset includes comprehensive, high-quality, human-provided captions that describe nearly every element within an image.
These detailed captions closely reflect the visual content, making them a reliable basis for evaluating the quality of long candidate captions.
However, the DCI dataset only provides pairs of images and references.
Because it lacks candidates and the corresponding human judgments, we cannot directly use this dataset for evaluating metrics.

Therefore, we collected long candidate captions generated by ten different models and gathered human judgments for each image–candidate pair.
For each image, we generated long candidates using ten representative MLLMs and image captioning models.
We employed the same prompts for each respective MLLM to generate these candidates.

Subsequently, each pair of candidates and images was independently assessed by human annotators from three perspectives: \textit{Desc.}, \textit{Rel.}, and \textit{Flu.}.
Following previous studies~\cite{gpt4, instructblip, internvl, llavanext, llava-1.5, mmgpt, qwen-vl, sharegpt4v, blip2, git}, the annotators assessed candidates on a five-point scale based on detailed guidelines.
To support the evaluation of \textit{Desc.}, we utilized SAM~\cite{sam} to generate object masks corresponding to image regions.
These object masks served as visual cues to assist the annotators in determining the necessary level of detail in their evaluations.

To align with the DCI’s split of the validation and test sets, we divided the \textsc{Vela} test set into two subsets: TestA and TestB. 
TestA comprises all images from the DCI dataset’s validation set, whereas TestB includes all images from the DCI dataset’s test set.
Further details on the benchmark and its construction process are provided in Appendix~\ref{sec:construction}.

\subsection{Quantitative Results}
\begin{table*}[t]
    \centering
    \normalsize

    \vspace{-3mm}
    \setlength{\tabcolsep}{10pt}
    \scalebox{0.75}{
    \begin{tabular}{
        >{\raggedright\arraybackslash}p{7mm}
        >{\raggedright\arraybackslash}p{37mm}
        >{\centering\arraybackslash}p{12mm}
        >{\centering\arraybackslash}p{12mm}
        >{\centering\arraybackslash}p{12mm}
        >{\centering\arraybackslash}p{12mm}
        >{\centering\arraybackslash}p{12mm}
        >{\centering\arraybackslash}p{12mm}
        >{\centering\arraybackslash}p{25mm}
    }

    \toprule
    \multirow{3}{*}{} & \multirow{3}{*}{\textbf{Metrics}}
        & \multicolumn{3}{c}{\textbf{TestA} [$\tau_c$] $\uparrow$} 
        & \multicolumn{3}{c}{\textbf{TestB }[$\tau_c$] $\uparrow$} 
        & \multirow{2}{*}{\begin{tabular}{@{}c@{}}
            \textbf{Inference time} \\
            {[ms] $\downarrow$}
            \end{tabular}} \\
        \cmidrule(lr){3-5} \cmidrule(lr){6-8}
        & & \textit{\textbf{Desc.}} & \textit{\textbf{Rel.}} & \textit{\textbf{Flu.}} 
        & \textit{\textbf{Desc.}} & \textit{\textbf{Rel.}} & \textit{\textbf{Flu.}} 
        &  \\
    \midrule

    \multirow{15}{*}{\rotatebox{90}{\footnotesize \textbf{Image captioning metrics}}} 
    & {BLEU}          & {28.6} & {2.4}  & {25.5} & {32.0} & {-10.1} & {-3.5} & {0.46} \\
    & {CIDEr}         & {-7.0} & {6.7}  & {4.4}  & {4.0}  & {-3.4}  & {1.9}  & {1.3} \\
    & {CLIP-S}        & {24.5} & {18.6} & {25.5} & {27.3} & {22.5}  & {24.5} & {26} \\
    & {CLIP-S$_{\textit{avg}}$}   & {-8.6} & {11.5} & {3.2}  & {12.8} & {27.5}  & {28.4} & {200} \\
    & {RefCLIP-S}     & {13.4} & {7.3}  & {9.5}  & {21.2} & {10.3}  & {10.9} & {33} \\
    & {PAC-S}         & {24.8} & {14.7} & {23.6} & {27.6} & {25.7}  & {23.0} & {48} \\
    & {PAC-S$_{\textit{avg}}$}    & {-7.4} & {14.6} & {6.2}  & {6.6}  & {29.2}  & {28.4} & {360} \\
    & {RefPAC-S}      & {22.6} & {19.1} & {24.9} & {40.7} & {29.2}  & {27.9} & {52} \\
    & {Polos}         & {28.5} & {18.1} & {30.6} & {41.1} & {22.4}  & {20.0} & {33} \\
    & {DENEB}         & {10.3} & {18.4} & {22.2} & {31.3} & {35.7}  & \underline{32.6} & {47} \\
    & {PAC-S++}       & {29.7} & {21.4} & {34.2} & {28.1} & {21.9}  & {21.1} & {36} \\
    & {PAC-S++$_{\textit{avg}}$} & {-7.2} & {19.4} & {6.0}  & {14.1} & {32.4}  & {30.3} & {270} \\
    & {RefPAC-S++}   & {25.4} & {23.3} & {28.9} & {40.3} & {22.2}  & {24.2} & {40} \\

    \midrule

    \multirow{5}{*}{\rotatebox{90}{\footnotesize \textbf{LLM-as-a-Judge}}} 
    & {FLEUR}               & {17.3} & {2.6}  & {0.5}  & {12.6} & {10.6} & {-3.1} & {1300} \\
    & {RefFLEUR}            & {21.3} & {10.3} & {7.2}  & {28.1} & {12.3} & {17.5} & {1400} \\
    & {G-VEval}             & {28.3} & {22.5} & {18.2} & {38.1} & {22.2} & {19.2} & {1800} \\
    & {GPT4o w/o references}    & \underline{54.1$\pm$1.0} & \underline{36.8$\pm$6.3} & {20.9$\pm$1.0} & {43.6$\pm$2.0} & \underline{37.3$\pm$3.4} & {25.2$\pm$1.0} & {1900} \\
    & {GPT4o w/ references}   & {47.0$\pm$1.1} & {26.2$\pm$2.2} & \underline{35.4$\pm$2.9} & \underline{46.9$\pm$2.6} & {30.4$\pm$2.3} & {25.1$\pm$4.3} & {2000} \\

    \midrule
    & {\textsc{Vela} (Ours)} & {\textbf{56.4$\pm$1.3}} & {\textbf{40.0$\pm$1.1}} & {\textbf{57.4$\pm$1.3}} & {\textbf{54.0$\pm$0.4}} & {\textbf{52.3$\pm$1.1}} & {\textbf{39.0$\pm$2.3}} & {260} \\ 
    \midrule \midrule

    {} & {Human performance} & {56.1} & {46.6} & {24.5} & {48.9} & {52.6} & {24.4} & {---} \\
    \bottomrule
    \end{tabular}
    }
    \caption{
    Quantitative comparison with baseline metrics. \textbf{Bold} font indicates the best results and \underline{underlined} font indicates the second best results. ViT-L/14 is used as the backbone for metrics that rely on CLIP~\cite{clip}.
    }
    \vspace{-3mm}
    \label{tab:quantitative}
\end{table*}

\label{quantitative}
Table~\ref{tab:quantitative} presents a quantitative comparison with baseline metrics on the TestA and TestB sets.  
Following previous research on image captioning metrics~\cite{pac-s, pac-spp, tong2025gveval, bridge, Zeng_2024hicescore}, we adopted Kendall’s $\tau_b$ and $\tau_c$ to evaluate the metrics. 
Due to space constraints, Table~\ref{tab:quantitative} only displays the results for Kendall’s $\tau_c$. Results for $\tau_b$ are provided in Appendix~\ref{sec:quantitative_taub}.  
Table~\ref{tab:quantitative} also compares the inference time per sample for each evaluation metric.

We adopted BLEU~\cite{bleu}, CIDEr~\cite{cider}, CLIP-S~\cite{clipscore}, RefCLIP-S~\cite{clipscore}, PAC-S~\cite{pac-s}, RefPAC-S~\cite{pac-s}, Polos~\cite{polos}, DENEB~\cite{deneb}, FLEUR~\cite{lee2024fleur}, RefFLEUR~\cite{lee2024fleur}, PAC-S++~\cite{pac-spp}, RefPAC-S++~\cite{pac-spp}, and G-VEval~\cite{tong2025gveval} as baselines because they are standard, representative metrics in the field of automatic evaluation for image captioning.

Metrics such as CLIP-S, PAC-S, and PAC-S++, which are based on CLIP~\cite{clip}, have a maximum input text length limited to 77 tokens by CLIP.  
Therefore, for a fair comparison, we followed the approach in \cite{yao2024hifi} and employed modified versions of CLIP-S, PAC-S, and PAC-S++ (called CLIP-S$_{\textit{avg}}$, PAC-S$_{\textit{avg}}$, and PAC-S++$_{\textit{avg}}$, respectively) as baselines alongside their original implementations.  
Specifically, these modified versions computed the cosine similarity between each sentence in the paragraph and the image, then outputted the final score by calculating the average of these scores. 

Furthermore, we evaluated the performance of GPT-4o under both reference-free and reference-based settings.  
For GPT-4o w/o references and GPT-4o w/ references, we utilized a modified version of the FLEUR~\cite{lee2024fleur} prompt, specifically tailored to assess the three aspects \textit{Desc.}, \textit{Rel.}, and \textit{Flu.}  
For both \textsc{Vela} and GPT-4o (with and without reference), we report the mean and standard deviation over five runs.

\paragraph{Correlation with human judgments.}
Table~\ref{tab:quantitative} demonstrates that our proposed metric achieved scores of 56.4, 40.0, and 57.4 for \textit{Desc.}, \textit{Rel.}, and \textit{Flu.} on the TestA set, and 54.0, 52.3, and 39.0 on the TestB set, respectively.

\textsc{Vela} outperformed both the reference-free and reference-based versions of GPT-4o on the TestA set by 2.3 points in \textit{Desc.}, 3.2 points in \textit{Rel.}, and 22.0 points in \textit{Flu.}  
Similarly, on the TestB set, \textsc{Vela} achieved improvements of 5.3 points in \textit{Desc.}, 1.7 points in \textit{Rel.}, and 15.6 points in \textit{Flu.}, compared with all baseline metrics.

The differences in $\tau_c$ between the proposed metric and each baseline metric were statistically significant (p < 0.05) for \textit{Rel.} and \textit{Flu.} on the TestA set, and for \textit{Desc.}, \textit{Rel.}, and \textit{Flu.} on the TestB set.

\paragraph{Inference time.}
Table~\ref{tab:quantitative} also shows the inference times per sample on our LongCap-Arena benchmark, evaluated using a GeForce RTX 3090 GPU and an Intel Core i9-10900KF CPU.
LLM-free metrics such as CLIP-S$_{\textit{avg}}$, PAC-S$_{\textit{avg}}$, and PAC-S++$_{\textit{avg}}$ demonstrated inference times ranging from approximately 1~ms to 400~ms.  
Moreover, existing LLM-based metrics such as FLEUR, RefFLEUR, G-VEval, and GPT-4o exhibited significantly longer inference times of 1280~ms, 1392~ms, 1812~ms, and 1905~ms, respectively—all values exceeding 1000~ms.  
In contrast, \textsc{Vela} achieved an inference time of only 258~ms, which is approximately five times faster than the LLM-based metrics.
These measurements include the time for both tokenization and CUDA kernel launches for a fair comparison.

\subsection{Qualitative Analysis}
\begin{figure*}[t]
    \centering
    \scalebox{1.0}{
        \includegraphics[width=\linewidth]{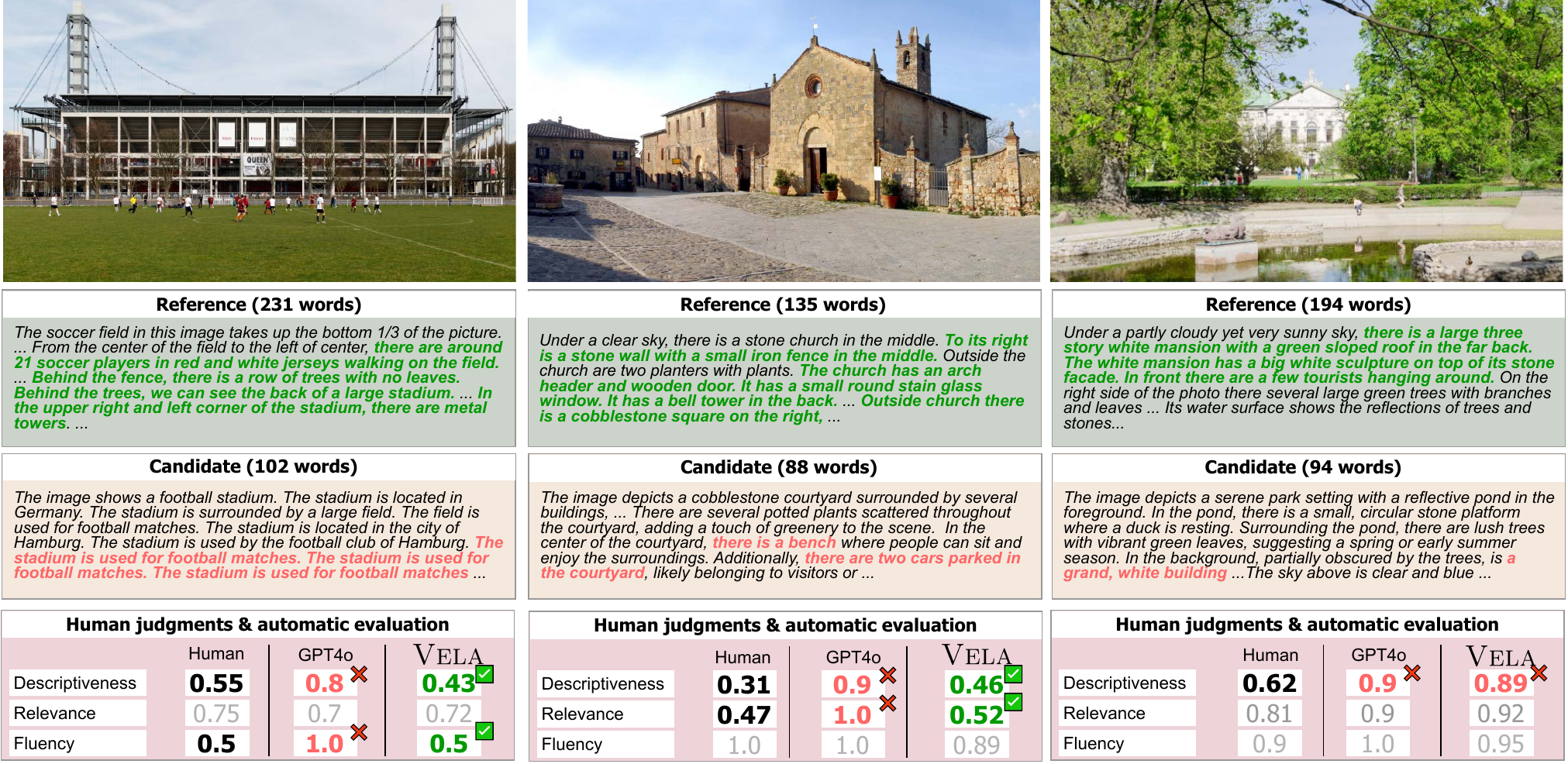}
    }
    \caption{Qualitative results on LongCap-Arena. The left and middle subfigures illustrate successful cases, while the right subfigure shows a failure case. Each subfigure consists of $\bm{x}_{\text{img}}$, $\bm{x}_{\text{ref}}$ (``Reference''), $\bm{x}_{\text{cand}}$ (``Candidate''), and human judgments $\bm{y}$ along with automatic evaluation scores $\hat{\bm{y}}$ (``Human judgments \& automatic evaluation''). Values in \textcolor{green!60!black}{green} and \textcolor{red}{red} indicate scores that are closely aligned and misaligned with human judgments, respectively.}
    \label{fig:qualitative}
\end{figure*}

In the example on the left, although $\bm{x}_{\text{cand}}$ captured the primary visible elements of the image, it lacked the detailed description found in $\bm{x}_{\text{ref}}$.
Therefore, the human annotators assigned it a \textit{Desc.} score of 0.55.  
Although GPT-4o w/o references overestimated the \textit{Desc.} score at 0.8, \textsc{Vela} evaluated it more appropriately with a score of 0.43, which aligns closely with the human judgment.
Note that $\bm{x}_{\text{cand}}$ contained repetitions (e.g., ``The stadium is used for football matches''), which led the human annotators to assign a \textit{Flu.} score of 0.5.  
In contrast, GPT-4o overestimated the \textit{Flu.} score, assigning a score of 1.0, whereas \textsc{Vela} evaluated it correctly, assigning a score of 0.5.

The middle subfigure presents another successful example for \textsc{Vela}.  
Here, $\bm{x}_{\text{cand}}$ primarily described the dominant elements of the image but failed to capture the details described in $\bm{x}_{\text{ref}}$; therefore, the human annotators assigned it a \textit{Desc.} score of 0.31.  
However, GPT-4o overestimated the \textsc{Desc.}, assigning a score of 0.9, whereas \textsc{Vela} evaluated it more appropriately at 0.46.
In the sample, $\bm{x}_{\text{cand}}$ included hallucinated objects (e.g., ``a bench'' and ``two cars parked in the courtyard''), leading human annotators to assign a \textit{Rel.} score of 0.47.  
GPT-4o failed to evaluate this correctly, assigning a score of 0.9.  
In contrast, \textsc{Vela} evaluated it more appropriately, assigning a score of 0.46.

In the right-hand subfigure of Fig.~\ref{fig:qualitative}, $\bm{x}_{\text{ref}}$ provided a detailed description of the white building at the center of the image, as indicated by the green text.  
Although the building occupied a relatively small region, this level of detail was likely motivated by its central placement and the absence of other semantically important objects.  
In contrast, $\bm{x}_{\text{cand}}$ described it only as ``a grand white building,'' which lacked the level of detail provided in $\bm{x}_{\text{ref}}$.
Given that the human judgment for \textit{Desc.} was 0.62, it is desirable for the automatic evaluation metrics not to overestimate the quality of this candidate.  
However, both the proposed metric and GPT-4o assigned inappropriately high scores for \textit{Desc.}, at 0.89 and 0.9, respectively.

To identify the cause, we examined the output of the R2C-LLM branch, without fusing it with the output of the I2C-Align branch.  
Although the fused output yielded a higher score of 0.89, the R2C-LLM branch outputted a score of 0.75, which was closer to the human judgment.  
This result suggests that the I2C-Align branch may have contributed to the discrepancy, possibly because it failed to recognize the white building as a key object.

\subsection{Ablation Studies}
\begin{table*}[t]
    \centering
    \normalsize
    \vspace{-3mm}
    \scalebox{0.78}{
    \begin{tabular}{
        >{\raggedright\arraybackslash}p{10mm}
        >{\centering\arraybackslash}p{22mm}
        >{\centering\arraybackslash}p{17mm}
        >{\centering\arraybackslash}p{40mm}
        >{\centering\arraybackslash}p{12mm}
        >{\centering\arraybackslash}p{12mm}
        >{\centering\arraybackslash}p{12mm}
        >{\centering\arraybackslash}p{12mm}
        >{\centering\arraybackslash}p{12mm}
        >{\centering\arraybackslash}p{12mm}
    }
    \toprule
    \multirow{2}{*}{\textbf{Metric}} 
    & \multirow{2}{*}{\begin{tabular}{@{}c@{}}\textbf{R2C-LLM} \\\textbf{backbone}\end{tabular}}
    & \multirow{2}{*}{\textbf{LLM-Hybrid}} 
    & \multirow{2}{*}{\begin{tabular}{@{}c@{}}\textbf{I2C-Align} \\\textbf{backbone}\end{tabular}}
    & \multicolumn{3}{c}{\textbf{TestA} [$\tau_c$] $\uparrow$} 
    & \multicolumn{3}{c}{\textbf{TestB} [$\tau_c$] $\uparrow$} \\
    \cmidrule(lr){5-7} \cmidrule(lr){8-10}
    {} & {} & {} & {} 
    & \textit{\textbf{Desc.}} & \textit{\textbf{Rel.}} & \textit{\textbf{Flu.}} 
    & \textit{\textbf{Desc.}} & \textit{\textbf{Rel.}} & \textit{\textbf{Flu.}} \\
    
    \midrule
    (i) & --- & --- & Long-CLIP ViT-L/14 & 45.9 & 11.9 & 11.6 & 24.6 & 28.8 & 4.9 \\
    (ii) & Qwen2.5-3B & --- & --- & 50.2 & 35.8 & 56.3 & 48.5 & 49.5 & 38.8 \\
    (iii) & Qwen2.5-3B & $\checkmark$ & CLIP ViT-L/14 & 54.9 & 37.2 & 51.5 & 50.0 & 46.7 & 34.5 \\
    (iv) & Qwen2.5-3B & $\checkmark$ & PAC-S CLIP ViT-L/14 & 54.4 & 37.7 & 53.1 & 51.9 & 48.0 & 32.1 \\
    (v) & Qwen2.5-3B & $\checkmark$ & Long-CLIP ViT-L/14 & 56.0 & 39.4 & 57.2 & 51.6 & 49.1 & 34.1 \\
    (vi) & Llama3.2-3B & $\checkmark$ & Long-CLIP ViT-L/14 & 54.5 & 36.2 & 51.8 & 51.0 & 49.8 & \textbf{41.3} \\
    (vii) & Phi-3.5-Mini & $\checkmark$ & Long-CLIP ViT-L/14 & 51.0 & 30.7 & 54.8 & 44.7 & 48.1 & 32.2 \\
    \midrule
    (viii) & Qwen2.5-3B & $\checkmark$ & Long-CLIP ViT-L/14 & \textbf{56.4$\pm$1.3} & \textbf{40.0$\pm$1.1} & \textbf{57.4$\pm$1.3} & \textbf{54.0$\pm$0.4} & \textbf{52.3$\pm$1.1} & 39.0$\pm$2.3 \\
    \bottomrule
    \end{tabular}
    }
    \caption{Results of ablation studies on the effect of incorporating the LLM-Hybrid-as-a-Judge framework and using different R2C-LLM and I2C-Align backbones. These results demonstrated that integrating the Long-CLIP ViT-L/14 backbone and LLM-Hybrid-as-a-Judge framework significantly contributed to the performance.}
    \vspace{-4mm}
    \label{tab:ablation_study}
\end{table*}

Table~\ref{tab:ablation_study} shows the quantitative results of the ablation studies. 
We conducted three ablation studies to investigate the contribution of each module in our proposed metric.

\noindent \textbf{\textit{LLM-Hybrid Ablation.}} 
We investigated the contribution of the LLM-Hybrid-as-a-Judge framework by excluding each branch. 
As shown in Table~\ref{tab:ablation_study}, a comparison between Metric (i) and Metric (viii) indicates that excluding the R2C-LLM branch led to decreases of 10.5, 28.1, and 45.8 points on TestA and 29.4, 23.5, and 34.1 points on TestB for \textit{Desc.}, \textit{Rel.}, and \textit{Flu.}, respectively. 
In contrast, Metric (ii), which excludes the I2C-Align branch, also showed performance decreases of 6.2, 4.2, and 1.1 points on TestA and 5.5, 2.8, and 0.5 points on TestB for \textit{Desc.}, \textit{Rel.}, and \textit{Flu.}, respectively.
These results demonstrate that both branches contribute to the performance improvement.

\noindent \textbf{\textit{I2C-Align Backbone Ablation.}} 
We analyzed the impact of different backbones in the I2C-Align branch by replacing the Long-CLIP ViT-L/14 backbone with alternative models. 
Table~\ref{tab:ablation_study} shows that Metric (viii) outperformed Metric (iii) with the CLIP ViT-L/14 backbone, Metric (iv) with the PAC-S CLIP ViT-L/14 backbone, and Metric (v) with the Long-CLIP ViT-B/16 backbone.

Specifically, Metric (viii) achieved improvements of 2.1, 6.3, and 6.9 points in \textit{Desc.}, \textit{Rel.}, and \textit{Flu.} on the TestB set compared with Metric (iv) using the PAC-S CLIP ViT-L/14 backbone, respectively.
These results demonstrate that the Long-CLIP ViT-L/14 backbone played a crucial role in enhancing performance.

\noindent \textbf{\textit{R2C-LLM Backbone Ablation.}} 
We investigated the effect of different R2C-LLM backbones by replacing the Qwen2.5-3B backbone with Llama3.2-3B and Phi-3.5 Mini.
We selected these models because they were lightweight yet high-performing, with model sizes comparable to Qwen2.5-3B.
Table~\ref{tab:ablation_study} shows that Metric (viii) outperformed Metric (v) with the Llama3.2-3B backbone and Metric (vii) with the Phi-3.5 Mini backbone.
Specifically, on the TestA set, Metric (viii) achieved improvements of 1.9, 3.8, and 5.6 points in \textit{Desc.}, \textit{Rel.}, and \textit{Flu.}, respectively, compared with the results of Metric (vi) using the Llama3.2-3B backbone, respectively.
These results demonstrate that the Qwen2.5-3B backbone played a crucial role in enhancing performance.

\subsection{Comparison with Human Performance}
We conducted a subject experiment to evaluate human performance on the TestA and TestB sets.  
Six participants participated in the experiment and were divided into two groups of three.  
Each group was assigned to evaluate the three aspects of long captions on either the TestA set or TestB set.

We calculated Kendall's $\tau_c$ for each evaluator's judgments against the ground truth in our dataset, and then computed the average Kendall's $\tau$ across all evaluators to measure the human performance.  
As shown in Table~\ref{tab:quantitative}, the human performance results for \textit{Desc.}, \textit{Rel.}, and \textit{Flu.} were 55.1, 41.0, and 28.1 on the TestA set, and 48.7, 50.6, and 23.4 for the TestB set, respectively.

Table~\ref{tab:quantitative} shows that the proposed metric outperformed human evaluation in both \textit{Desc.} and \textit{Flu.}
Specifically, the proposed metric outperformed human evaluation by 0.3 and 5.1 points in \textit{Desc.}, and by 32.9 and 14.6 points in \textit{Flu.}, on the TestA and TestB sets respectively.
These results indicate that the proposed metric achieved performance comparable to human evaluation in assessing the \textit{Desc.} of candidates.
Moreover, the large margin in \textit{Flu.} suggests that the proposed metric could potentially replace human evaluation in assessing naturalness and grammatical correctness.

By contrast, Table~\ref{tab:quantitative} shows that in \textit{Rel.}, the proposed method underperformed human performance by 6.6 points on the TestA set and 0.3 points on the TestB set.
As discussed in Appendix~\ref{sec:error_analysis}, this performance gap could be attributed to insufficient grounding in the I2C-Align branch and the suboptimal integration of outputs from the R2C-LLM and I2C-Align branches.
A possible solution to this is to integrate visual information while leveraging a pretrained language model, similar to the gated xattn-dense layer in Flamingo~\cite{flamingo}.

\section{Conclusion}
In this study, we focused on the automatic evaluation of long and detailed image captions generated by MLLMs.
The contributions of this study are as follows:
(i) We proposed \textsc{Vela}, a supervised metric evaluating long image captions from three distinct perspectives.
(ii) We introduced the LLM-Hybrid-as-a-Judge framework, which enables computationally efficient and LLM-based evaluations while incorporating images through the R2C-LLM and I2C-Align branches.
(iii) We constructed LongCap-Arena, a benchmark designed for both training and evaluating metrics on long captions, featuring 32,246 human judgments collected from 1,020 annotators.
(iv) \textsc{Vela} outperformed existing metrics and achieved superhuman performance on the LongCap-Arena benchmark.

\section{Limitations}
Although our metric has clearly been shown to provide a high correlation with human judgments, it is not without its limitations. The primary limitation is the occurrence of errors stemming from the lack of sufficient detail or accuracy in the references. Moreover, the metric tends to erroneously overlook semantically important objects in the image, especially when they occupy relatively small regions. These limitations could be attributed to insufficient grounding in the I2C-Align branch and the suboptimal integration of outputs from the R2C-LLM and I2C-Align branches. 
Another important limitation is that the R2C-LLM branch requires access to last hidden states, which prevents the direct use of closed-source models.
For further error analysis, see Appendix~\ref{sec:error_analysis}

\section*{Acknowledgments}
This work was supported by a grant from Apple Inc. Any views, opinions, findings, and conclusions or recommendations expressed in this material are those of the authors and should not be interpreted as reflecting the views, policies, or position, either expressed or implied, of Apple Inc.
This work was also partially supported by JSPS KAKENHI Grant Number 23K28168 and JST Moonshot.

\bibliography{reference}
\newpage
\appendix
\section{Additional Related Work}
\label{sec:additional_related_work}
CLIP-S~\cite{clipscore} and PAC-S~\cite{pac-s} use CLIP~\cite{clip} to compute the cosine similarity between the vector representations of the image and candidate.
In contrast, Polos~\cite{polos} and DENEB~\cite{deneb} employ supervised learning based on human judgments, achieving high correlation with human judgments.
CLAIR~\cite{chan2023clair} is an LLM-based metric that uses GPT-3.5 to evaluate candidates with respect to human-provided references. 
However, it does not incorporate visual information, which limits its applicability to image-grounded tasks.

\section{Construction of LongCap-Arena}
\label{sec:construction}
\begin{figure*}[t]
    \centering
    \includegraphics[width=0.95\linewidth]{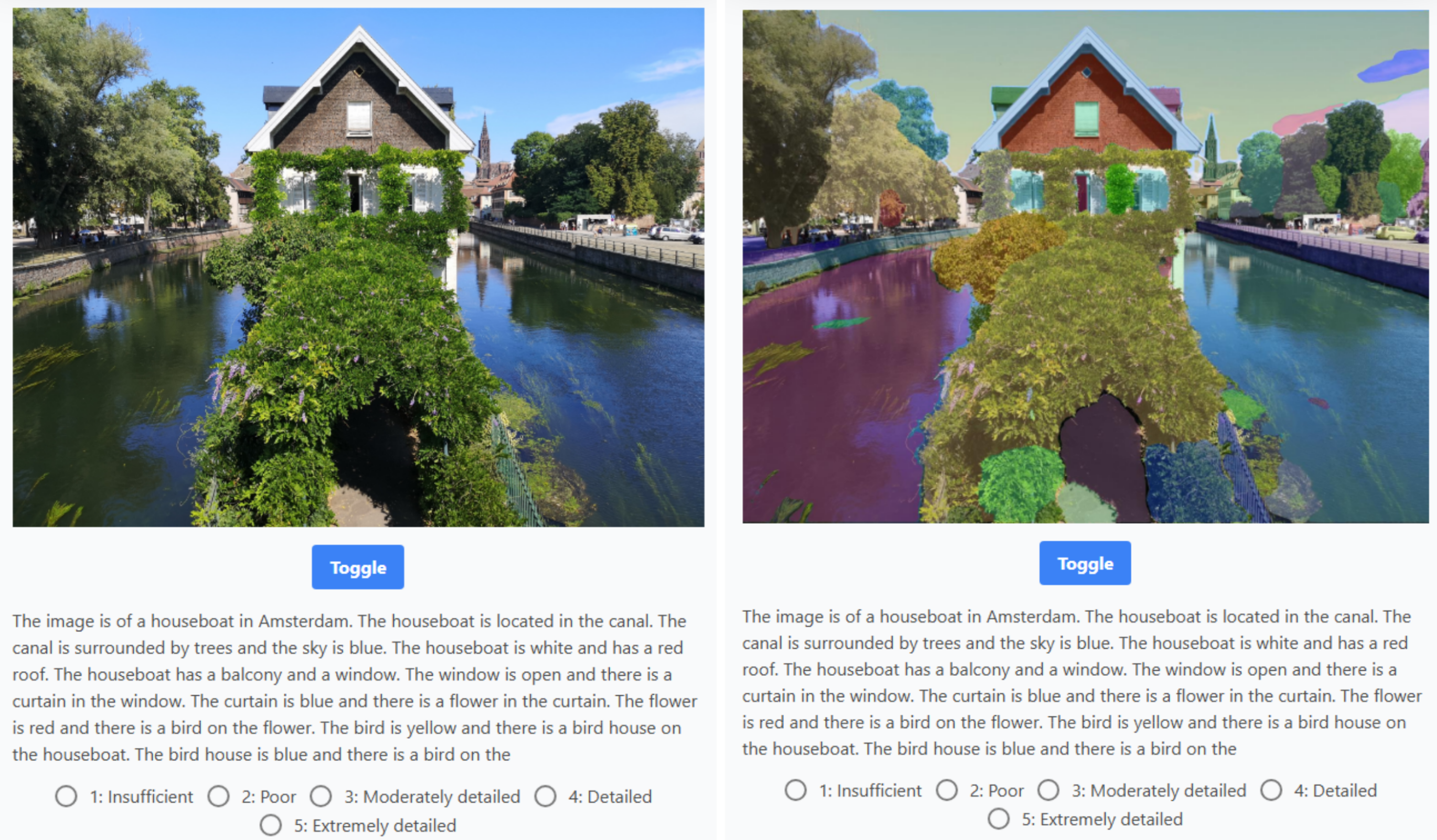}
    \caption{Annotation interface for \textit{Desc}. The left subfigure shows the normal image, and the right subfigure presents its segmented version generated using SAM~\cite{sam}. These object masks were shown to annotators as visual cues, helping them determine the level of detail required for their evaluation.}
    \label{fig:annotation}
\end{figure*}

In this study, we constructed LongCap-Arena, a new benchmark designed for evaluating metrics for long captions.
LongCap-Arena contains 32,246 human judgments and 7,805 images, each paired with human-provided long reference captions, and long candidate captions.
The candidate captions were generated by ten representative MLLMs and image captioning models based on the DCI dataset images.
These models include GPT-4o~\cite{gpt4}, InstructBLIP~\cite{instructblip}, InternVL~\cite{internvl}, LLaVA-NeXT~\cite{llavanext}, LLaVA-1.5~\cite{llava-1.5}, MultimodalGPT~\cite{mmgpt}, Qwen-VL-Chat~\cite{qwen-vl}, ShareGPT4V~\cite{sharegpt4v}, BLIP2~\cite{blip2}, and GIT~\cite{git}. 

The candidate captions included 7,805 captions with a vocabulary size of 21,611 words, a total word count of 570,600, and an average length of 101.2 words.
Moreover, the reference captions consisted of 7,805 captions with a vocabulary size of 20,988 words, a total word count of 738,848, and an average length of 131.4 words. 
Moreover, all captions are in English.

This dataset is built on the DCI dataset~\cite{dci}, which provides images and long reference captions annotated by humans.
When creating a new dataset from existing images in the DCI dataset, it is crucial to carefully address the issue of potential data leakage in MLLMs, which could arise if the training set of the DCI dataset is also used to train the MLLMs.
Therefore, we constructed the training and validation sets of the \textsc{Vela} dataset using the training set of the DCI dataset.

The training, validation, TestA, and TestB sets contain 11,971, 1,309, 294, and 324 samples, respectively.
The training set was used for training the metric, the validation set for hyperparameter tuning, and the test set for evaluating the metric's performance.

Human judgments were provided on a five-point scale and subsequently normalized to the range $[0,1]$. 
Each candidate caption was evaluated by a minimum of three distinct annotators. 
The final score for each caption was determined by calculating the average of these individual judgments.
Fig.~\ref{fig:annotation} illustrates the annotation interface used for evaluating \textit{Desc.}
The annotation was conducted via a public crowdsourcing platform, where we recruited annotators from a general population on the internet without restricting demographic or geographic background.
We recruited annotators and provided payment that was adequate based on the participants’ country of residence, and obtained consent via the task instructions, which clearly stated that the collected data would be used for research purposes.
To ensure data reliability, we excluded responses exhibiting suspicious behavior, such as excessively short response times or repeated identical scores.

\section{Implementation Details}
\begin{table}[h]
    \normalsize
    \newcommand*{\bhline}[1]{\noalign{\hrule height #1}}
    \centering
    \resizebox{\linewidth}{!}{
    \begin{tabular}{lc}
        \toprule
        Epoch & $10$ \\ \hline
        Optimizer & AdamW ($\beta_{1}=0.9$, $\beta_{2}=0.999$) \\ \hline
        Learning rate & $1.0 \times 10^{-4}$ \\ \hline
        Batch size & $4$ \\
        \bottomrule
    \end{tabular}
    }
    \caption{Settings of the proposed metric.}
    \label{tab:settings}
\end{table}

Table~\ref{tab:settings} shows the training settings of the proposed metric.
Our metric had approximately 3.68 million trainable parameters.
Our metric was trained on a system equipped with an NVIDIA GeForce RTX 3090 GPU with 24GB memory and an Intel Core i9-12900K CPU with 64GB RAM.
The training process was completed within approximately three hours. 
We employed early stopping during training using Kendall's $\tau_c$.  
Specifically, $\tau_c$ was computed on the validation set after each epoch.  
Training was terminated when no improvement in $\tau_c$ was observed on the validation set for a single epoch.  
Subsequently, the metric's performance was assessed on the test set.

\section{Quantitative Results for Kendall's $\tau_b$}
\label{sec:quantitative_taub}
\begin{table*}[t]
    \centering
    \normalsize
    \setlength{\tabcolsep}{8pt}
    \begin{tabular}{lcccccc}
        \toprule
        \textbf{Metrics} & \multicolumn{3}{c}{\textbf{TestA} [$\tau_b$] $\uparrow$} & \multicolumn{3}{c}{\textbf{TestB} [$\tau_b$] $\uparrow$} \\
        \cmidrule(lr){2-4} \cmidrule(lr){5-7}
        & \textit{\textbf{Desc.}} & \textit{\textbf{Rel.}} & \textit{\textbf{Flu.}} & \textit{\textbf{Desc.}} & \textit{\textbf{Rel.}} & \textit{\textbf{Flu.}} \\
        \midrule
        GPT4o w/o references & 56.1$\pm$1.3 & 38.2$\pm$6.2 & 35.2$\pm$1.0 & 44.6$\pm$1.8 & 38.8$\pm$4.1 & \textbf{39.3$\pm$3.0} \\
        GPT4o w/ references & 47.4$\pm$1.1 & 27.3$\pm$2.3 & 37.0$\pm$3.2 & 47.2$\pm$2.6 & 31.7$\pm$2.3 & 27.1$\pm$5.3 \\
        \midrule
        \textsc{Vela} (Ours) & \textbf{56.8$\pm$1.1} & \textbf{41.0$\pm$1.2} & \textbf{55.2$\pm$0.9} & \textbf{52.2$\pm$1.8} & \textbf{51.9$\pm$1.2} & 38.7$\pm$2.1 \\
        \midrule \midrule
        Human performance & 54.4 & 47.5 & 32.1 & 46.6 & 51.9 & 31.1 \\
        \bottomrule
    \end{tabular}
    \caption{
    Quantitative results (Kendall's $\tau_b$) on the LongCap-Arena benchmark. \textsc{Vela} outperforms GPT-4o (w/ and w/o references) across \textit{Desc.}, \textit{Rel.}, and \textit{Flu.}, and notably surpasses human performance in \textit{Desc.} and \textit{Flu.} 
    }
    \label{tab:quantitative_supp}
\end{table*}

Table~\ref{tab:quantitative_supp} presents additional results using Kendall’s $\tau_b$, confirming that \textsc{Vela} achieved superior performance compared with GPT-4o (with and without references) on both the TestA and TestB sets.

\section{Error Analysis}
\label{sec:error_analysis}
To investigate the limitations of the proposed metric, we analyzed samples on which the proposed metric did not perform as expected.  
We defined failure cases for each evaluation perspective as samples where the corresponding output satisfied the following condition:
\[
|y_{\text{per}} - \hat{y}_{\text{per}}| \geq \theta, \quad \text{per} \in \{\text{desc}, \text{rel}, \text{flu}\}
\]
In this study, we fixed $\theta$ at a value of 0.25, as this value corresponds to the difference between two adjacent points on a normalized five-point scale.  
Under this condition, we identified 9, 11, and 5 failure cases in \textit{Desc.}, \textit{Rel.}, and \textit{Flu.}, respectively, from the combined TestA and TestB sets, which comprise 206 samples in total.

\begin{table}[t]
    \centering
    \normalsize
    \vspace{2mm}
    \begin{tabular}{
        >{\raggedright\arraybackslash}p{60mm}
        >{\centering\arraybackslash}p{10mm}
    }
    \toprule
    \textbf{Failure Mode Category} & \#Errors \\
    \midrule    
    Insufficient detail or accuracy in references & 12 \\
    Redundant candidates & 2 \\
    Over-reliance on references & 3 \\
    Named entities in candidates & 1 \\
    Fluency issues & 3 \\
    Short candidates & 2 \\
    Others & 2 \\
    \midrule
    \textbf{Total} & 25 \\
    \bottomrule
    \end{tabular}
    \caption{Categorization of failure modes.}
    \label{tab:error_table}
\end{table}

Table~\ref{tab:error_table} shows the categorization of failure modes, which are grouped into the following categories:
\begin{itemize}[leftmargin=10pt,itemsep=0pt]
  \item \underline{Insufficient detail or accuracy in references}:\\
  This category encompasses failure modes where the references lack sufficient image-related details, leading to evaluation discrepancies in \textit{Desc.}
  \item \underline{Redundant candidates}:\\
  This category refers to failure modes where the proposed metric has assigned inappropriate scores to candidates with redundant information unrelated to the image.
  \item \underline{Incorrect or missing information in references}:\\
  This category pertains to failure modes where the references contain incorrect or missing information, leading to evaluation discrepancies in \textit{Rel.}
  \item \underline{Over-reliance on references}:\\
  This category refers to failure modes where the proposed metric has prioritized the references over the image, leading to evaluation scores that differ from human judgments.
  \item \underline{Named entities in candidates}:\\
  This category refers to failure modes where candidates include named entities whose correctness cannot be determined solely based on the image, leading to evaluation discrepancies.
  \item \underline{Fluency issues}:\\
  This category encompasses failure modes where the proposed metric has assigned scores that do not align with human judgments to candidates containing unnatural elements (e.g., unnatural phrasing, grammatical or spelling errors, redundancy, or extraneous characters).
  \item \underline{Short candidates}:\\
  This category refers to failure modes where the proposed metric has assigned inappropriate scores to short candidates because the training data predominantly consisted of long captions.
  \item \underline{Others}:\\
  This category encompasses various errors that do not fall into the aforementioned categories.
\end{itemize}
We independently categorized failure modes in each evaluation perspective into the above categories.
Table~\ref{tab:error_table} shows that the primary cause of errors was the lack of sufficient detail or accuracy in the references.
These errors likely arise because the proposed metric does not effectively handle the outputs from the I2C-Align branch, which do not rely on references, and fails to adequately integrate the outputs of the R2C-LLM and I2C-Align branches.
In future work, we plan to extend our metric by introducing a mechanism that integrates visual information while leveraging a pretrained language model, similar to the gated xattn-dense layer in Flamingo~\cite{flamingo}.

\section{Fusion of Visual and Textual Features}
\label{sec:fusion}

\begin{table*}[t]
    \centering
    \normalsize
    \setlength{\tabcolsep}{8pt}
    \begin{tabular}{lcccccc}
        \toprule
        \textbf{Metrics} & \multicolumn{3}{c}{\textbf{TestA} [$\tau_c$] $\uparrow$} & \multicolumn{3}{c}{\textbf{TestB} [$\tau_c$] $\uparrow$} \\
        \cmidrule(lr){2-4} \cmidrule(lr){5-7}
        & \textit{\textbf{Desc.}} & \textit{\textbf{Rel.}} & \textit{\textbf{Flu.}} & \textit{\textbf{Desc.}} & \textit{\textbf{Rel.}} & \textit{\textbf{Flu.}} \\
        \midrule
        \textsc{Vela} w/ Transformer & 53.3 & 39.5 & 57.6 & 52.0 & 52.3 & 37.0 \\
        \textsc{Vela} w/ MLP (Ours) & \textbf{56.4$\pm$1.3} & \textbf{40.0$\pm$1.1} & \textbf{57.4$\pm$1.3} & \textbf{54.0$\pm$0.4} & \textbf{52.3$\pm$1.1} & \textbf{39.0$\pm$2.3} \\
        \bottomrule
    \end{tabular}
    \caption{Comparison of Transformer-based and MLP-based fusion between visual and textual features. While the Transformer-based fusion achieved competitive results, the MLP-based fusion demonstrated better performance in \textit{Desc.}}
    \label{tab:fusion_study}
\end{table*}
To investigate potential limitations in expressivity, we conducted experiments on the comparison of Transformer-based and MLP-based fusion between visual and textual features.
Table~\ref{tab:fusion_study} shows the results of the variant modified to employ Transformer-based fusion.
The results demonstrate that the Transformer-based fusion achieved performance comparable to the MLP-based fusion overall, with slightly lower scores in \textit{Desc}.
Given these results and the computational cost, we adopted MLP-based fusion in the final model.

\section{\textsc{Vela} in Reference-free Setting}
\label{sec:ref-free}

\begin{table*}[t]
    \centering
    \normalsize
    \setlength{\tabcolsep}{5pt}
    \begin{tabular}{lcccccc}
        \toprule
        \textbf{Metrics} & \multicolumn{3}{c}{\textbf{TestA} [$\tau_c$] $\uparrow$} & \multicolumn{3}{c}{\textbf{TestB} [$\tau_c$] $\uparrow$} \\
        \cmidrule(lr){2-4} \cmidrule(lr){5-7}
        & \textit{\textbf{Desc.}} & \textit{\textbf{Rel.}} & \textit{\textbf{Flu.}} & \textit{\textbf{Desc.}} & \textit{\textbf{Rel.}} & \textit{\textbf{Flu.}} \\
        \midrule
        GPT-4o w/o references & 54.1$\pm$1.0 & 36.8$\pm$6.3 & 20.9$\pm$1.0 & 43.6$\pm$2.0 & 37.3$\pm$3.4 & 25.2$\pm$1.0 \\
        GPT-4o w/ references  & 47.0$\pm$1.1 & 26.2$\pm$2.2 & 35.4$\pm$2.9 & 46.9$\pm$2.6 & 30.4$\pm$2.3 & 25.1$\pm$4.3 \\
        \midrule
        \textsc{Vela} w/o references & 54.3 & 36.1 & 56.4 & 46.4 & 44.4 & 33.4 \\
        \textsc{Vela} w/ references (Ours) & \textbf{56.4$\pm$1.3} & \textbf{40.0$\pm$1.1} & \textbf{57.4$\pm$1.3} & \textbf{54.0$\pm$0.4} & \textbf{52.3$\pm$1.1} & \textbf{39.0$\pm$2.3} \\
        \bottomrule
    \end{tabular}
    \caption{Quantitative results of \textsc{Vela} in a reference-free setting. Reference-free \textsc{Vela} outperformed GPT-4o in both reference-free and reference-based settings.}
    \label{tab:ref_free}
\end{table*}
\textsc{Vela} can be used in a reference-free setting by simply removing the reference input from the R2C-LLM prompt, without any modification to the architecture. 
Table~\ref{tab:ref_free} shows the results of evaluating \textsc{Vela} in the reference-free setting.
Although the absence of human-annotated references led to a decrease in performance compared to the reference-based \textsc{Vela}, reference-free \textsc{Vela} outperformed GPT-4o in both settings. 
Specifically, reference-free \textsc{Vela} demonstrates an improvement over reference-free GPT-4o by +0.3, -0.7, and +35.5 points on TestA, and by +2.8, +19.2, and +8.2 points on TestB, in \textit{Desc.}, \textit{Rel.}, and \textit{Flu.}, respectively. 
These results demonstrate that \textsc{Vela} can generalize well to real-world scenarios where reference captions are unavailable.

\section{Zero-shot Evaluation on Short Caption Benchmarks}
\label{sec:zeroshot}

\begin{table*}[h]
    \centering
    \normalsize
    \setlength{\tabcolsep}{8pt}
    \begin{tabular}{lcc}
        \toprule
        \textbf{Metric} & \textbf{Composite [$\tau_c$] $\uparrow$} & \textbf{Flickr8k-Expert [$\tau_c$] $\uparrow$} \\
        \midrule
        CIDEr \cite{cider}        & 37.7 & 43.9 \\
        CLIP-S \cite{clipscore}   & 53.8 & 51.2 \\
        RefCLIP-S \cite{clipscore} & 55.4 & 53.0 \\
        PAC-S \cite{pac-s}         & 55.7 & 54.3 \\
        RefPAC-S \cite{pac-s}      & 57.3 & 55.9 \\
        Polos \cite{polos}        & 57.6 & \underline{56.4} \\
        CLAIR \cite{chan2023clair}        &  --  & 48.8 \\
        FLEUR \cite{lee2024fleur}        & \textbf{63.5} & 53.0 \\
        G-VEval \cite{tong2025gveval}     &  --  & \textbf{59.7} \\
        \midrule
        \textsc{Vela} (Ours)      & \underline{61.3} & 56.2 \\
        \bottomrule
    \end{tabular}
    \caption{Results of zero-shot evaluation on the short caption benchmarks, Composite and Flickr8k-Expert. \textsc{Vela} achieved comparable performance with existing state-of-the-art metrics.}
    \label{tab:zeroshot}
\end{table*}
To evaluate potential overfitting to the DCI dataset~\cite{dci}, we additionally conducted experiments in a zero-shot setting on two standard benchmarks for short caption evaluation: Composite~\cite{composite} and Flickr8k-Expert~\cite{flickr}.
In these experiments, we modified \textsc{Vela} to output a single score instead of three perspective scores, following the evaluation protocol of these benchmarks.
Table~\ref{tab:zeroshot} shows that \textsc{Vela} achieved competitive performance with existing state-of-the-art metrics~\cite{cider, clipscore, pac-s, polos, chan2023clair, lee2024fleur, tong2025gveval} on both benchmarks.  
These results indicate that \textsc{Vela} was not overfitted to the DCI dataset and remains effective for short caption evaluation.

\section{Scoring Criteria for Annotation}
\paragraph{Descriptiveness:} \mbox{}\\
Annotators were instructed to evaluate how detailed the caption was in describing the image content, focusing on objects, relationships, and attributes. \\
They used both the \textit{``normal image''} and \textit{``segmented image''} (with color-coded objects) to assess the level of detail. \\
The scoring criteria were as follows:
\begin{itemize}
    \setlength{\parskip}{0.5mm}
    \setlength{\itemsep}{0.2mm}
    \item[5:] \underline{Extremely detailed} — The caption comprehensively describes all observed objects and relationships, including spatial relationships and contextual details.
    \item[4:] \underline{Detailed} — The caption describes most objects and relationships, with only minor omissions.
    \item[3:] \underline{Moderately detailed} — The caption mentions key objects but lacks detail in relationships or other attributes.
    \item[2:] \underline{Poor} — The caption includes descriptions of a few objects but omits significant details and relationships.
    \item[1:] \underline{Insufficient} — The caption provides minimal or no description of the image content.
\end{itemize}

\paragraph{Relevance:} \mbox{}\\
Relevance was evaluated based on the correctness of objects, attributes, and relationships mentioned in the captions. \\
Proper nouns and associated specific details (e.g., \textit{``Mt. Fuji''}) were excluded from the evaluation. \\
The scoring criteria were as follows:
\begin{itemize}
    \setlength{\parskip}{0.5mm}
    \setlength{\itemsep}{0.2mm}
    \item[5:] \underline{Fully relevant} — The caption appropriately describes the image content without errors.
    \item[4:] \underline{Mostly relevant} — Minor inaccuracies are present but the overall caption is almost correct.
    \item[3:] \underline{Partially relevant} — Significant inaccuracies exist, yet some parts of the caption remain correct.
    \item[2:] \underline{Barely relevant} — Numerous inaccuracies significantly distort the relevance of the caption.
    \item[1:] \underline{Not relevant} — The caption is fundamentally unrelated to the image content.
\end{itemize}

\paragraph{Fluency:} \mbox{}\\
Annotators were directed to evaluate the naturalness and grammatical correctness of captions, independent of their accuracy. \\
Markdown syntax (e.g., \verb|###|, \verb|-|) was not considered an error. \\
The scoring criteria were as follows:
\begin{itemize}
    \setlength{\parskip}{0.5mm}
    \setlength{\itemsep}{0.2mm}
    \item[5:] \underline{Extremely fluent} — No errors or minimal errors (no more than one).
    \item[4:] \underline{Fluent} — Minor errors present but the caption is generally natural and comprehensible.
    \item[3:] \underline{Moderately fluent} — Noticeable errors are present but the text remains understandable.
    \item[2:] \underline{Lacking fluency} — Numerous errors make the caption difficult to read.
    \item[1:] \underline{Not fluent} — Frequent errors render the caption incomprehensible.
\end{itemize}

\section{Prompts in \textsc{Vela}}
\label{sec:prompts}

This section provides the full prompts used in the R2C-LLM branch of \textsc{Vela} for \textit{Desc.}, \textit{Rel.}, and \textit{Flu.}

\subsection{Descriptiveness}
\noindent\rule{\linewidth}{0.5pt}
\noindent\texttt{System}\\
Evaluate the descriptiveness of the candidate caption based on the reference captions and the provided image. Focus only on how detailed the caption is, regardless of relevance. Refer to the following criteria:\\
- 5: Extremely detailed - Captures all objects, relationships, and attributes in the image with precise and complete descriptions, including spatial relationships and overall context.\\
- 4: Detailed - Captures most objects and relationships but lacks some elements.\\
- 3: Partially detailed - Mentions key objects but misses spatial relationships or additional details.\\
- 2: Insufficient detail - Mentions only a few objects correctly, with many elements missing.\\
- 1: Very poor detail - Mentions almost no objects and fails to represent the image content.\\
Only give a number from 1 to 5 with no text.\\
\\
\texttt{User}\\
Reference Captions: \{\{Reference\}\}\\
Candidate Caption: \{\{Candidate\}\}\\
\\
\texttt{Assistant}\\
Score:

\noindent\rule{\linewidth}{0.5pt}

\subsection{Relevance}
\noindent\rule{\linewidth}{0.5pt}

\noindent\texttt{System}\\
Evaluate the relevance of the candidate caption to the provided image considering the reference captions. Focus solely on how well the caption aligns with the image content, ignoring fluency or descriptiveness. Refer to the following criteria:\\
- 5: Fully relevant - Accurately describes the image content with no errors.\\
- 4: Mostly relevant - Contains minor errors but is generally aligned with the image content.\\
- 3: Partially relevant - Includes significant errors but some parts relate to the image.\\
- 2: Barely relevant - Contains many errors and deviates significantly from the image content.\\
- 1: Not relevant - Contains numerous errors and fundamentally mismatches the image.\\
Only give a number from 1 to 5 with no text.\\
\texttt{User}\\
\\
Reference Captions: \{\{Reference\}\}\\
Candidate Caption: \{\{Candidate\}\}\\
\\
\texttt{Assistant}\\
Score:

\noindent\rule{\linewidth}{0.5pt}

\subsection{Fluency}

\noindent\rule{\linewidth}{0.5pt}
\noindent\texttt{System}\\
Evaluate the fluency of the candidate caption, focusing solely on its grammatical correctness, naturalness, and readability. Ignore the content's relevance or descriptiveness. Refer to the following criteria:\\
- 5: Very fluent - No errors or only one minor error; reads naturally as proper English sentences.\\
- 4: Fluent - Contains some errors but is overall natural and easy to understand.\\
- 3: Partially fluent - Noticeable errors but still comprehensible.\\
- 2: Lacking fluency - Many errors that make it hard to read.\\
- 1: Not fluent - Excessive errors that make it incomprehensible.\\
Only give a number from 1 to 5 with no text.\\
\\
\texttt{User}\\                                                                                         
Reference Captions: \{\{Reference\}\}\\
Candidate Caption: \{\{Candidate\}\}\\
\\
\texttt{Assistant}\\
Score:

\noindent\rule{\linewidth}{0.5pt}

\section{Additional Details for ARR Checklist}

\paragraph{Discuss The License For Artifacts}
\textsc{Vela} and LongCap-Arena are released under the BSD 3-Clause Clear License.

The licenses of the models and datasets used in this study are summarized below:

\noindent
\begin{tabularx}{\linewidth}{@{}X@{}}
DCI dataset~\cite{dci}: \\[0.1em] \hfill CC BY-NC 4.0 \\
InstructBLIP~\cite{instructblip}: \\[0.1em] \hfill Research (non-commercial) \\
InternVL~\cite{internvl}: \\[0.1em] \hfill MIT license \\
LLaVA-NeXT~\cite{llavanext}: \\[0.1em] \hfill Apache 2.0 license \\
LLaVA-1.5~\cite{llava-1.5}: \\[0.1em] \hfill Apache 2.0 license \\
Multimodal-GPT~\cite{mmgpt}: \\[0.1em] \hfill Apache 2.0 license \\
Qwen-VL-Chat~\cite{qwen-vl}: \\[0.1em] \hfill Tongyi Qianwen License \\
ShareGPT4V~\cite{sharegpt4v}: \\[0.1em] \hfill Apache 2.0 license \\
BLIP2~\cite{blip2}: \\[0.1em] \hfill BSD 3-Clause license \\
GIT~\cite{git}: \\[0.1em] \hfill MIT license \\
Long-CLIP~\cite{long-clip}: \\[0.1em] \hfill Apache 2.0 license \\
Qwen2.5~\cite{qwen2.5}: \\[0.1em] \hfill Apache 2.0 license \\
\end{tabularx}

\paragraph{Artifact Use Consistent With Intended Use}
All existing artifacts used in this study were utilized in a manner consistent with their intended use. For the artifacts we created, we define their intended use as general academic and research use, which is compatible with the original access conditions of the datasets and models employed in this study.

\paragraph{Data Contains Personally Identifying Info Or Offensive Content}
The collected data contain no personally identifiable or offensive content. 
All data used in this study are publicly available. 
We also confirmed that the source websites, repositories, and publications include no statements indicating concerns about personal information.
\end{document}